# Multi-view Subspace Clustering via An Adaptive Consensus Graph Filter

Lai Wei
weilai@shmtu.edu.cn
College of Information Engineering, Shanghai Maritime University
Shanghai, P.R. China

Shanshan Song
ssssong163@163.com
College of Information Engineering, Shanghai Maritime University
Shanghai, P.R. China

## ABSTRACT

Multiview subspace clustering (MVSC) has attracted an increasing amount of attention in recent years. Most existing MVSC methods first collect complementary information from different views and consequently derive a consensus reconstruction coefficient matrix to indicate the subspace structure of a multi-view data set. In this paper, we initially assume the existence of a consensus reconstruction coefficient matrix and then use it to build a consensus graph filter. In each view, the filter is employed for smoothing the data and designing a regularizer for the reconstruction coefficient matrix. Finally, the obtained reconstruction coefficient matrices from different views are used to create constraints for the consensus reconstruction coefficient matrix. Therefore, in the proposed method, the consensus reconstruction coefficient matrix, the consensus graph filter, and the reconstruction coefficient matrices from different views are interdependent. We provide an optimization algorithm to obtain their optimal values. Extensive experiments on diverse multi-view data sets demonstrate that our approach outperforms some state-of-the-art methods.

**ACM Reference Format:**
Lai Wei and Shanshan Song. 2024. Multi-view Subspace Clustering via An Adaptive Consensus Graph Filter. In . ACM, New York, NY, USA, 9 pages. https://doi.org/XXXXXXX.XXXXXXX

## 1 INTRODUCTION

As information technology advances, data about an object can be gathered from multiple sources. For instance, a document can be described in different languages, and a grayscale image can be characterized by various features such as scale-invariant feature transform (SIFT) [6], histogram of oriented gradients (HOG) [5], local binary patterns (LBP) [25] and so on. Since the features from different views of the same object provide complementary information, it is important to develop data processing methods for multi-view data.

Subspace clustering arranges the high-dimensional data samples into a union of linear subspaces where they are generated from [1, 26], then the subspace structure of the data set is recovered. Over the past few decades, a range of subspace clustering algorithms, including sparse subspace clustering (SSC) [7], low-rank representation (LRR) [20], and least square regression (LSR) [21] have been proposed. These methods have gotten promising achievements in many real applications such as face clustering [15, 27] and motion segmentation [38].

In recent years, as previously mentioned, with the increasing volume of emerging multi-view data, subspace clustering algorithms have been extended to the multi-view learning domain. Various multi-view subspace clustering (MVSC) algorithms [2, 3, 13, 14, 16, 23, 24, 30–32] that aim to utilize complementary information within multi-view data sets have been proposed. For example, Chen et al. project data points from different views into a common latent space and consequently seek a reconstruction coefficient matrix to represent the intrinsic subspace structure of the data set [3]. Zhang et al. also explore the latent representation of multi-view variables to discover underlying complementary information which is beneficial for LRR [36]. Cao et al. minimize the Hilbert-Schmidt Independence Criterion (HSIC) of the reconstruction coefficient matrices obtained in different views to get consistent structure information of a multi-view data set [2]. Li et al. apply HSIC to learn a consensus representation of a multi-view data set and employ LRR to get a coefficient matrix by using the learned representation [16]. Along this line, Wang et al. define a position-aware exclusivity term to measure the diversity of representations obtained from different views [32].

In addition to the aforementioned algorithms, some advanced techniques such as deep learning [4, 35], tensorization [30], and graph filtering [9, 18] have been introduced into the field of multi-view subspace clustering. In particular, graph filtering-based methods have demonstrated both its simplicity and effectiveness in solving multi-view subspace clustering problems. For instance, building upon the graph filtering-based subspace clustering algorithm presented in [22], Huang et al. put forward a graph filtering-based MVSC approach known as LSRMVSC (Learning Smooth Representation for Multi-View Subspace Clustering) [12] and its variant in [11]. In this paper, we devise a new multi-view subspace clustering method leveraging the graph filtering technique. In contrast to LSRMVSC, which constructs a distinct graph filter for each view of data, our approach involves the creation of a consensus graph filter aimed at acquiring smooth data representations across all views. In the proposed algorithm, this consensus graph filter is derived from a consensus reconstruction coefficient matrix, consolidating consistent information in the reconstruction coefficient matrices obtained from different views. Meanwhile, in each view, by using the consensus graph filter, we design a new regularizer for the reconstruction coefficient matrix obtained in this view. Therefore, the consensus graph filter, the consensus reconstruction coefficient matrix, and the reconstruction coefficient matrices acquired from different views are interdependent. We present an optimization algorithm to update these variables iteratively and show that the optimal solutions could be finally achieved.

The main contributions of this paper could be summarized as follows:

1. We propose to use a consensus low-pass graph filter to smooth the feature and devise a regularizer for the reconstruction coefficient matrix sought in each view of a multi-view data set.

2. We present a joint optimization problem that integrates the computation of reconstruction coefficient matrices from different views and the extraction of information from these reconstruction coefficient matrices to build a consensus reconstruction matrix.

3. We develop an iterative optimization algorithm to adaptively learn the consensus reconstruction coefficient matrix and the reconstruction coefficient matrix in each view.

## 2 PRELIMINARIES
### 2.1 Multi-view subspace clustering

The generalized framework of MVSC could be expressed as follows:

$$\min_{\mathbf{C}^i} \left\{ \gamma^i \left[ \mathcal{L}(\mathbf{X}^i, \mathbf{C}^i \mathbf{X}^i) + \lambda \mathcal{R}(\mathbf{C}^i) \right] \right\}_{i=1}^{v}, \quad (1)$$

where $\mathbf{X}^i \in \mathbb{R}^{n \times d^i}$ and $\mathbf{C}^i \in \mathbb{R}^{n \times n}$ is the feature matrix and the reconstruction coefficient matrix in the $i$-th view. $n$ is the number of samples, $d^i$ is the number of features in the $i$-th view and $v$ is the number of views. $\mathcal{L}(\cdot, \cdot)$ is the loss functions that measures how well $\mathbf{X}^i$ is approximated by $\mathbf{C}^i \mathbf{X}^i$. $\mathcal{R}$ is a regularizer used to help $\mathbf{C}^i$ to discover the subspace structure of $\mathbf{X}^i$ or even explore the relationships of multiple views. $\gamma^i$ characterizes the importance of $i$-th view and $\lambda > 0$ is a parameter.

Once the reconstruction coefficient matrix $\mathbf{C}^i$ in each view is obtained, and then a consensus matrix will be defined as $\mathbf{C} = \mathcal{F}(\{\mathbf{C}^i\}_{i=1}^{v})$, where $\mathcal{F}(\cdot)$ is some kind fusion function. Finally, a spectral clustering algorithm, e.g. normalized cuts (Ncuts) [29] is performed on an affinity matrix $\mathbf{W} = (|\mathbf{C}| + |\mathbf{C}^\top|)/2$ to produce the final clustering results, where $\mathbf{C}^\top$ is the transpose of $\mathbf{C}$.

### 2.2 Graph filter

Suppose an undirected graph $\mathbb{G}$ has $n$ vertices with $\mathbf{X}$ being the feature matrix corresponding to these vertices. And $\mathbf{W} \in \mathbb{R}^{n \times n}$ represents the edge weights matrix with $[\mathbf{W}]_{ij} = [\mathbf{W}]_{ji} \geq 0$, $[\mathbf{W}]_{ij}$ indicates the $(i,j)$–th element of $\mathbf{W}$. Then the normalized adjacency matrix and the normalized Laplacian matrix of this graph are defined as $\tilde{\mathbf{A}} = \mathbf{W} + \mathbf{I}$ and $\mathbf{L} = \mathbf{I} - \tilde{\mathbf{D}}^{-\frac{1}{2}} \tilde{\mathbf{A}} \tilde{\mathbf{D}}^{-\frac{1}{2}}$ respectively, where $\mathbf{I} \in \mathbb{R}^{n \times n}$ is an identity matrix, $\tilde{\mathbf{D}}$ is a diagonal matrix and $[\tilde{\mathbf{D}}]_{ii} = \sum_{j=1}^{n} [\tilde{\mathbf{A}}]_{ij}$.

$\mathbf{L}$ could be eigen-decomposed as $\mathbf{L} = \mathbf{U} \mathbf{\Lambda} \mathbf{U}^\top$, where $\mathbf{U} = [\mathbf{u}_1, \cdots, \mathbf{u}_n]$, $\mathbf{\Lambda} = Diag(\lambda_1, \cdots, \lambda_n)$ are eigenvalue matrix and eigenvalues are sorted in increasing order, $Diag(\cdot)$ reformulates a vector to be the diagonal of a matrix. The set of eigenvectors of $\mathbf{L}$ can be considered as the graph's Fourier basis and the eigenvalues as the associated frequencies. Then a graph signal could be represented as a linear combination of the eigenvectors, i.e., $\mathbf{y} = \sum_{i=1}^{n} t_i \mathbf{u}_i = \mathbf{U} \mathbf{t}$ and $\mathbf{t} = [t_1; \cdots; t_n]$ is a coefficient vector. As stated in [37], the smoothness of a basis signal $\mathbf{u}_k$ can be measured as $\mathbf{u}_k^\top \mathbf{L} \mathbf{u}_k = \lambda_k, 1 \leq k \leq n$. Hence, the smaller the eigenvalue, the smoother the basis graph signal. Consequently, graph signals consisting of basis signals with smaller eigenvalues (low-frequencies) will be smooth.

Thus for a graph signal $\mathbf{y}$, we can get its smoothed signal $\bar{\mathbf{y}}$ as:

$$\bar{\mathbf{y}} = \mathbf{G} \mathbf{y} = \left( \mathbf{U} p(\mathbf{\Lambda}) \mathbf{U}^\top \right) \cdot (\mathbf{U} \mathbf{t}) = \sum_{i=1}^{n} p(\lambda_i) t_i \mathbf{u}_i. \quad (2)$$

where $p(\lambda_i)$ is a low-pass response function. Since the eigenvalues of a symmetric normalized graph Laplacian fall into $[0, 2]$, we can define $p(\lambda_i) = 1 - \frac{\lambda_i}{2}$ [19, 37], then the low-pass graph filter $\mathbf{G} = \mathbf{I} - \mathbf{L}/2$. It can be seen that in the filtered signal $\bar{\mathbf{y}}$, the coefficient $t_i$ of the basis signal $\mathbf{u}_i$ is scaled by $p(\lambda_i)$. Therefore, $\bar{\mathbf{y}}$ will be more smooth.

## 3 THE PROPOSED METHOD
### 3.1 Motivation

The proposed method contains three main steps: creation of a consensus graph filter, computation of the reconstruction coefficient matrix in each view, and amalgamation of the reconstruction coefficient matrices from different views to obtain a consensus coefficient matrix. The overall pipeline of the proposed method can be visualized in Fig. 1. We then describe the steps in detail.

**A. Designing a consensus graph filter.** As we know, the fundamental goal of MVSC is to obtain a consensus coefficient matrix that consolidates consistent information from different views. To introduce our method, we first assume that a consensus coefficient matrix $\mathbf{C}$ from a multi-view data set $\mathbf{X} = \{\mathbf{X}^i\}_{i=1}^{v}$ has been gotten. Then from the subspace clustering viewpoint, a weight affinity matrix could be obtained as $\mathbf{W} = (|\mathbf{C}| + |\mathbf{C}^\top|)/2$. If we force $\mathbf{C} = \mathbf{C}^\top$ and $\mathbf{C} \geq 0$, then $\mathbf{W} = \mathbf{C}$. Consequently, considering the graph filter technique, if we further compel $\mathbf{C} \mathbf{1} = \mathbf{1}$ and $diag(\mathbf{C}) = \mathbf{0}$, the normalized Laplacian matrix $\mathbf{L} = \mathbf{I} - \tilde{\mathbf{D}}^{-\frac{1}{2}} \tilde{\mathbf{A}} \tilde{\mathbf{D}}^{-\frac{1}{2}} = (\mathbf{I} - \mathbf{C})/2$. Then based on the description in the previous section, we can define a low-pass graph filter $\mathbf{G} = \mathbf{I} - \mathbf{L}/2 = \frac{3}{4} \mathbf{I} + \frac{1}{4} \mathbf{C}$.

**B. Computing the reconstruction coefficient matrix in each view.** Given that $\mathbf{C}$ is a consensus matrix, the defined graph filter $\mathbf{G}$ should be consistent across different views. Hence, the smoothed features of a multi-view data set in each view could be computed, i.e., in the $i$-th view, the smoothed feature matrix $\mathbf{Y}^i = \mathbf{G} \mathbf{X}^i$. Meanwhile, by following the methodology of subspace clustering, we could hope $\mathbf{Y}^i$ has the self-expressive property, namely, $\mathbf{Y}^i \approx \mathbf{C}^i \mathbf{Y}^i$, where $\mathbf{C}^i$ is the reconstruction coefficient matrix. Because $\mathbf{C}^i$ is always viewed as a new representation of $\mathbf{Y}^i$ [7, 20], $\mathbf{G} \mathbf{C}^i$ serves as the smooth representation $\mathbf{C}^i$. Suppose $\mathbf{C}^i$ could reveal the subspace structure of $\mathbf{X}^i$, $\mathbf{G} \mathbf{C}^i$ should have the same ability. This implies $\mathbf{C}^i$ and $\mathbf{G} \mathbf{C}^i$ should be very similar which leads to a new constraint of $\mathbf{C}^i$ as $\|\mathbf{C}^i - \mathbf{G}\mathbf{C}^i\|_F^2$. This technique was first proposed in [33], however different from [33], here $\mathbf{G}$ is a consensus graph filter and $\mathbf{C}^i$ is the reconstruction coefficient matrix in the $i$-th view. By combining the self-expressive term, we could define the following subspace clustering problem in $i$-th view:

$$\begin{aligned}\min_{\mathbf{C}^i} \quad & \|\mathbf{Y}^i - \mathbf{C}^i\mathbf{Y}^i\|_F^2 + \alpha\|\mathbf{C}^i - \mathbf{G}\mathbf{C}^i\|_F^2,\\ s.t. \quad & \mathbf{Y}^i = \mathbf{G}\mathbf{X}^i.\end{aligned} \quad (3)$$

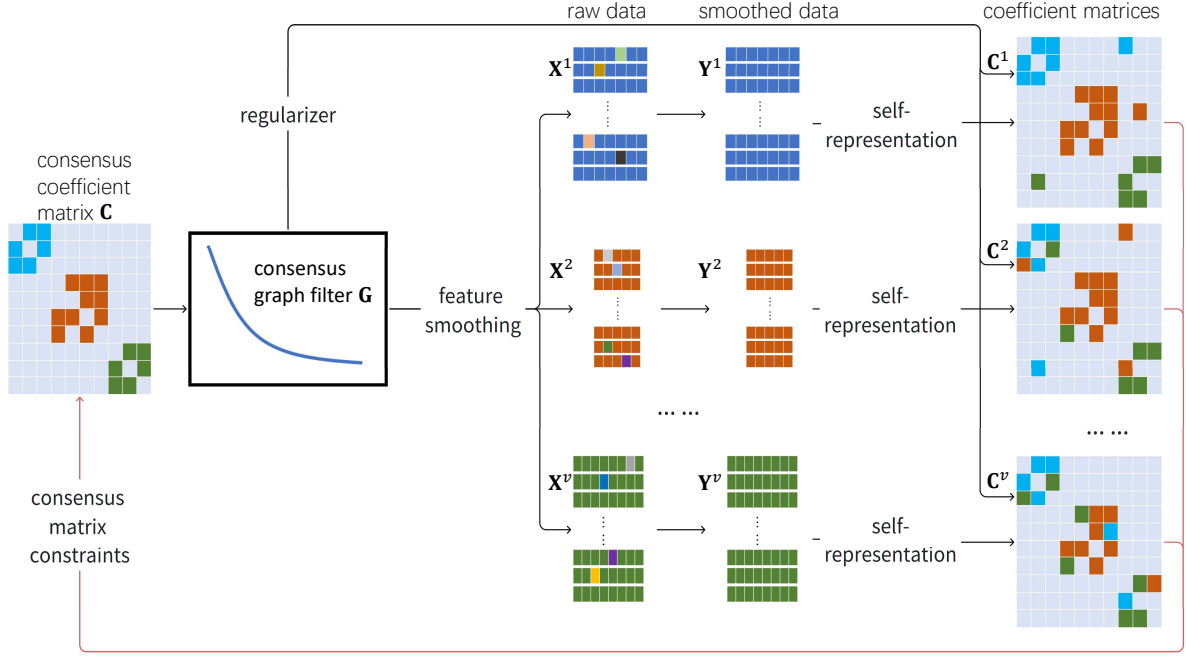

**Figure 1: The overview of the proposed method.** The consensus graph filter $\mathbf{G}$ is used to obtain the smoothed data for each view. Meanwhile, in each view, a regularizer of the coefficient matrix is devised by $\mathbf{G}$. Then a coefficient matrix could be obtained in each view. These coefficient matrices are used to construct constraints for obtaining the consensus coefficient matrix $\mathbf{C}$. Finally, $\mathbf{G}$ is derived by $\mathbf{C}$.

Substitute $\mathbf{G} = \frac{3}{4}\mathbf{I} + \frac{1}{4}\mathbf{C}$ into the above function and through a simple derivation, the following problem could be obtained:

$$\begin{aligned}\min_{\mathbf{C}^i} \quad & \|\mathbf{Y}^i - \mathbf{C}^i\mathbf{Y}^i\|_F^2 + \alpha\|\mathbf{C}^i - \mathbf{C}\mathbf{C}^i\|_F^2,\\ s.t. \quad & 4\mathbf{Y}^i = 3\mathbf{X}^i + \mathbf{C}\mathbf{X}^i.\end{aligned} \quad (4)$$

**C. Fusing the reconstruction coefficient matrices.** We discuss how to get the consensus coefficient matrix $\mathbf{C}$. A straightforward assumption is that the consensus coefficient matrix should closely approximate the coefficient matrices obtained in different views. Hence, a constraint term as $\sum_{i=1}^v \|\mathbf{C} - \mathbf{C}^i\|_F^2$ could be easily built. Considering the importance of different views, this term can be refined as $\sum_{i=1}^v (\gamma^i)^\eta \|\mathbf{C} - \mathbf{C}^i\|_F^2$, where $\eta$ is a parameter. Then by integrating Problem (4), we could define the following MVSC problem:

$$\begin{aligned}\min_{\mathbf{Y}^i,\mathbf{C}^i,\mathbf{Z}^i,\mathbf{C},\gamma^i} \quad & \sum_{i=1}^v \|\mathbf{Y}^i - \mathbf{C}^i\mathbf{Y}^i\|_F^2 + \alpha\|\mathbf{C}^i - \mathbf{C}\mathbf{C}^i\|_F^2\\ & +\beta(\gamma^i)^\eta\|\mathbf{C} - \mathbf{C}^i\|_F^2,\\ s.t. \quad & 4\mathbf{Y}^i = 3\mathbf{X}^i + \mathbf{C}\mathbf{X}^i, \mathbf{C}^i\mathbf{1} = \mathbf{1},\\ & \mathbf{C}^i = (\mathbf{C}^i)^\top, \mathbf{C}^i \geq 0, diag(\mathbf{C}^i) = \mathbf{0},\\ & \sum_{i=1}^v \gamma^i = 1, \gamma^i > 0, i = 1,\cdots,v,\\ & \mathbf{C}\mathbf{1} = \mathbf{1}, \mathbf{C} = \mathbf{C}^\top, \mathbf{C} \geq 0, diag(\mathbf{C}) = \mathbf{0},\end{aligned} \quad (5)$$

where $\alpha, \beta$ are two positive hyper-parameters. The constraints on $\mathbf{C}$ originate from the priori assumptions we derived for this problem. We apply the same constraints to $\mathbf{C}^i$ with the aim of making the obtained $\mathbf{C}^i$ approximate $\mathbf{C}$. Problem (5) is termed Multi-View Subspace Clustering by an adaptive Consensus Graph Filter (MVSC$^2$GF).

### 3.2 Optimization
We use ADMM (alternating direction method of multipliers method) [17] to solve Problem (5). Firstly, its equivalent formulation could

be expressed as follows:

$$\begin{aligned}
\min_{Y^i, C^i, Z^i, C, Z, \gamma^i} \quad & \sum_{i=1}^{v} \|Y^i - C^i Y^i\|_F^2 + \alpha \|C^i - CZ^i\|_F^2 \\
& + \beta(\gamma^i)^\eta \|C - C^i\|_F^2, \\
s.t. \quad & 4Y^i = 3X^i + CX^i, C^i \mathbf{1} = \mathbf{1}, C^i = Z^i \\
& Z^i = (Z^i)^\top, Z^i \geq 0, diag(Z^i) = 0, \\
& \sum_{i=1}^{v} \gamma^i = 1, \gamma^i > 0, i = 1, \cdots, v, \\
& C\mathbf{1} = \mathbf{1}, C = Z, \\
& Z = Z^\top, Z \geq 0, diag(Z) = 0,
\end{aligned} \quad (6)$$

where $Z^i (i = 1, 2, \cdots, v)$ and $Z$ are auxiliary variables. Then the augmented Lagrangian function of Problem (6) is:

$$\begin{aligned}
\mathbb{L}_1 =& \sum_{i=1}^{v} \Big[ \|Y^i - C^i Y^i\|_F^2 + \alpha \|C^i - CZ^i\|_F^2 \\
& + \beta(\gamma^i)^\eta \|C - C^i\|_F^2 + <\Gamma^i, 4Y^i - 3X^i - CX^i> \\
& + <\Lambda^i, C^i - Z^i> + <\Omega^i, C^i \mathbf{1} - \mathbf{1}> \\
& + \mu/2 \big( \|4Y^i - 3X^i - CX^i\|_F^2 + \|C^i - Z^i\|_F^2 \\
& \|C^i \mathbf{1} - \mathbf{1}\|_F^2 \big) \Big] + <\Theta, C - Z> + <\Phi, C\mathbf{1} - \mathbf{1}> \\
& + \mu/2 \big( \|C - Z\|_F^2 + \|C\mathbf{1} - \mathbf{1}\| \big)
\end{aligned} \quad (7)$$

where $\{\Gamma^i, \Lambda^i, \Omega^i\}_{i=1}^{v}$, $\Theta$ and $\Phi$ are Lagrange multipliers, $\mu > 0$ is a parameter. By minimizing $\mathbb{L}_1$, the variables $\{Y^i, C^i, Z^i\}_{i=1}^{v}, C, Z$ and the weight $\{\gamma^i\}_{i=1}^{v}$ can be optimized alternately.

**1. Updating** $Y^i$. It is clear that $Y^i, C^i, Z^i$ can be updated separately for each view. The optimization Eq. (7) w.r.t $Y^i$ can be reduced to

$$\min_{Y^i} \|Y^i - C^i Y^i\|_F^2 + \mu/2 \|4Y^i - 3X^i - CX^i + \Gamma^i/\mu\|_F^2. \quad (8)$$

Taking the derivative of Eq. (8) with respect to $Y^i$, the analytical solution for $Y^i$ can be derived by

$$Y^i = \Big[ 2(I - C^i)^\top (I - C^i) + 16\mu I \Big]^{-1} \Big( 12\mu X^i + 4\mu C X^i - 4\Gamma^i \Big) \quad (9)$$

**2. Updating** $C^i$. The optimization Eq. (7) w.r.t $C^i$ can be turned to

$$\begin{aligned}
\min_{C^i} \quad & \|Y^i - C^i Y^i\|_F^2 + \alpha \|C^i - CZ^i\|_F^2 \\
& + \beta(\gamma^i)^\eta \|C^i - C\|_F^2 + \mu/2 \Big( \|C^i - Z^i + \Lambda^i/\mu\|_F^2 \\
& + \|C^i \mathbf{1} - \mathbf{1} + \Omega^i/\mu\|_F^2 \Big).
\end{aligned} \quad (10)$$

We take the derivative of Eq. (10) with respect to $C^i$, then the following equation holds:

$$\begin{aligned}
C^i = & \Big[ 2Y^i(Y^i)^\top + 2\big(\alpha CZ^i + \beta(\gamma^i)^\eta C\big) + \mu(Z^i + \mathbf{1}\mathbf{1}^\top) \\
& -\Lambda^i - \Omega^i \mathbf{1}^\top \Big] \Big[ 2Y^i(Y^i)^\top + 2\big(\alpha + \beta(\gamma^i)^\eta\big) I + \mu(I + \mathbf{1}\mathbf{1}^\top) \Big]^{-1}.
\end{aligned} \quad (11)$$

**3. Updating** $Z^i$. The optimization for $Z^i$ is

$$\min_{Z^i} \alpha \|C^i - CZ^i\|_F^2 + \mu/2 \|C^i - Z^i + \Lambda^i/\mu\|_F^2. \quad (12)$$

Then the updating scheme of $Z^i$ is

$$Z^i = \Big( 2\alpha C^\top C + \mu I \Big)^{-1} \Big( 2\alpha C^\top C^i + \mu C^i + \Lambda^i \Big). \quad (13)$$

After $Z^i$ is updated, we further let $Z^i = [Z^i + (Z^i)^\top]/2$, $Z^i = \max(Z^i, 0)$ and $diag(Z^i) = 0$, so that $Z^i$ satisfies the additional constraints.

**4. Updating** C. We could get the following optimization problem w.r.t. C by following the above methodology:

$$\begin{aligned}
\min_{C} \quad & \sum_{i=1}^{v} \Big[ \alpha \|C^i - CZ^i\|_F^2 + \beta(\gamma^i)^\eta \|C - C^i\|_F^2 \\
& + \mu/2 \|4Y^i - 3X^i - CX^i + \Gamma^i/\mu\|_F^2 \Big] \\
& + \mu/2 \Big( \|C - Z + \Theta/\mu\|_F^2 + \|C\mathbf{1} - \mathbf{1} + \Phi/\mu\| \Big).
\end{aligned} \quad (14)$$

Then the closed form solution of C is

$$C = AB^{-1}, \quad (15)$$

where $A = \sum_{i=1}^{v} \Big[ 2\alpha C^i(Z^i)^\top + 2\beta(\gamma^i)^\eta C^i + 4\mu Y^i(X^i)^\top - 3\mu X^i(X^i)^\top + \Gamma^i(X^i)^\top \Big] + \mu(Z + \mathbf{1}\mathbf{1}^\top) - \Theta - \Phi \mathbf{1}^\top$ and $B = \sum_{i=1}^{v} \Big[ 2\alpha Z^i(Z^i)^\top + 2\beta(\gamma^i)^\eta I + \mu X^i(X^i)^\top \Big] + \mu(I + \mathbf{1}\mathbf{1}^\top)$.

**5. Updating** Z. The optimization for Z could be simply expressed as follows:

$$\min_{Z} \|C - Z + \Theta/\mu\|_F^2. \quad (16)$$

Hence, we have

$$Z = C + \Theta/\mu. \quad (17)$$

Similar to the process for updating $Z^i$, we also make $Z = [Z + (Z)^\top]/2$, $Z = \max(Z, 0)$ and $diag(Z) = 0$.

**6. Updating Lagrange multipliers and** $\mu$. The precise updating schemes for Lagrange multipliers and $\mu$ are presented in Eq. (18):

$$\begin{cases}
\Gamma^i = \Gamma^i + \mu(4Y^i - 3X^i - CX^i), \\
\Lambda^i = \Lambda^i + \mu(C^i - Z^i), \\
\Omega^i = \Omega^i + \mu(C^i \mathbf{1} - \mathbf{1}), i = 1, 2 \cdots, v \\
\Theta = \Theta + \mu(C - Z), \\
\Phi = \Phi + \mu(C\mathbf{1} - \mathbf{1}), \\
\mu = \min(\mu_{max}, \rho\mu)
\end{cases} \quad (18)$$

where $\mu_{max}$ is a relative large value and $\rho$ is a positive scalar.

**7. Updating** $\gamma$. After the above variables have been updated, minimizing Eq. (7) can be simplified as

$$\begin{aligned}
\min_{\gamma^i} \quad & \sum_{i=1}^{v} (\gamma^i)^\eta J^i, \\
s.t. \quad & \sum_{i=1}^{v} \gamma^i = 1, \gamma^i > 0,
\end{aligned} \quad (19)$$

where $J^i = \|C - C^i\|_F^2$. The Lagrange function of Eq. (19) is

$$\mathbb{L}_2 = \sum_{i=1}^{v} (\gamma^i)^\eta J^i - \tau \Big( \sum_{i=1}^{v} \gamma^i - 1 \Big), \quad (20)$$

where $\tau$ is the Lagrange multiplier. By taking the derivative of $\mathbb{L}_2$ w.r.t. $\gamma^i$ to zero, we have

$$\gamma^i = \frac{(J^i)^{\frac{1}{1-\eta}}}{\sum_{i=1}^{v} (J^i)^{\frac{1}{1-\eta}}}. \quad (21)$$

### 3.3 Algorithm

We summarize the algorithmic procedure for solving MVSC$^2$GF problem (i.e., Eq. (6)) in Algorithm 1.

**Algorithm 1** Multi-view subspace clustering with an adaptive consensus graph filter

**Input:** The data matrix $\mathbf{X} = [\mathbf{x}_1; \mathbf{x}_2; \cdots; \mathbf{x}_n]$, three parameters $\alpha, \beta, \eta > 0$, the maximal number of iteration $Maxiter < 1e4$;
**Output:** The consensus coefficient matrix $\mathbf{C}$.
1: Initialize the parameters, i.e., $\mu = 10^{-6}$, $\mu_{max} = 10^{30}$, $\rho = 1.1$, $\varepsilon = 10^{-4}$ and $\mathbf{\Gamma}^i = \mathbf{Y}^i = \mathbf{0}, \mathbf{\Lambda}^i = \mathbf{C}^i = \mathbf{Z}^i = \mathbf{\Theta} = \mathbf{C} = \mathbf{Z} = \mathbf{0}, \mathbf{\Omega}^i = \mathbf{\Phi} = \mathbf{0}, \gamma^i = 1/v, i = 1, 2, \cdots, v$.
2: **while** not converged **do**
3:  Update variables $\mathbf{Y}^i, \mathbf{C}^i, \mathbf{Z}^i, \mathbf{C}$ and $\mathbf{Z}$ by using Eq.(9), Eq.(11),Eq.(13), Eq.(15) and Eq.(17) respectively;
4:  Update Lagrange multipliers $\mathbf{\Gamma}^i, \mathbf{\Lambda}^i, \mathbf{\Omega}^i, \mathbf{\Theta}, \mathbf{\Phi}$ and parameter $\mu$ by using Eq. (18);
5:  Update the weights $\gamma^i$ by using Eq. (21).
6: **end while**

### 3.4 Convergence and complexity analysis

We first analyze the convergence of Algorithm 1. The convergence of ADMM with two variables has been well established in [17]. But the number of variables that need to be optimized in Problem (6) is much greater than 2. Fortunately, it could be clearly seen that all the terms in the objective function of Problem (7) are strongly convex to the relative variables. Then according to Theorem 1 in [8], if $\mu$ is set less than a certain value, the variable sequence generated by ADMM will converge to a stationary solution to Problem (6). In the following experiment section, we show that Algorithm 1 converges in finite steps on all data sets.

Secondly, the computation time of Algorithm 1 mainly relies on the updating of variables $\{\mathbf{Y}^i, \mathbf{C}^i, \mathbf{Z}^i\}_{i=1}^v, \mathbf{C}$, and $\mathbf{Z}$. We can see all the variables have closed-form updating equations. Suppose the number of samples is $n$, it takes $O(n^3)$ to compute the pseudo-inverse of an $n \times n$ matrix and the multiplication of two $n \times n$ matrices for updating $\{\mathbf{Y}^i, \mathbf{C}^i, \mathbf{Z}^i\}_{i=1}^v$ and $\mathbf{C}$. And the complexity for updating $\mathbf{Z}$ is $O(n^2)$, hence the time complexity of Algorithm 1 in each iteration is $O(n^3)$. In our experiments, the iterations of Algorithm 1 are always less than 500, hence its total complexity is $O(n^3)$, which aligns with the complexity of some existing MVSC algorithms [2, 3, 16, 32].

## 4 EXPERIMENTS

In this section, we evaluate the clustering performance of MVSC$^2$GF alongside ten state-of-the-art multi-view subspace clustering methods on seven benchmark data sets[1]. Four widely-used metrics including clustering accuracy (ACC), normalized mutual information (NMI), adjusted rand index (ARI), and F-score are employed.

### 4.1 Experiment setup

**A. data set description**: The brief information on the seven used data sets is introduced as follows:

1. 3Sources[2]: This data set contains 169 articles that are manually sorted into six categories. Each article is collected from three online news sources, and each news source is treated as a single view of an article.

2. ORL[3]: It has 400 face images, distributed among 40 different subjects, with ten images for each subject. Three types of features describe each image.

3. MSRC-v1 [34]: This data set contains 210 objects in images belonging to seven categories. Each image is represented by five views.

4. BBCSport[4]: This is a textual data set containing 544 documents originating from the BBC Sport website. It is characterized by two views.

5. COIL-20[5]: This data set contains 1440 images with 20 classes. Each category includes 72 black background images of different locations. Three features are extracted as views.

6. Caltech101-7[6]: It includes 1474 images from Caltech101 with seven categories. This data set consists of six views.

7. Handwritten numerals (HW)[7]: It is a data set consisting of 2,000 images for "0" to "9" digit classes, 200 samples for each class. Each image is represented by six views of features.

**B. Compared Methods**: The ten representative multi-view subspace clustering methods (with the years in which they are proposed) used for comparison are listed as follows: diversity-induced multi-view subspace clustering (DiMSC, 2015 [2]), auto-weighted multiple graph learning (AMGL, 2016 [24]), multi-view learning with adaptive neighbors (MLAN, 2016 [23]), multiple partitions aligned clustering (mPAC, 2019 [13]), graph-based multi-view clustering (GMC, 2020 [31]), latent multi-view subspace clustering (LMSC, 2020 [36]), large-scale multi-view subspace clustering in linear time (LMVSC, 2020 [14]), learning smooth representation for multi-View subspace clustering (LSRMVSC, 2022 [12]), consistency-induced multi-view subspace clustering (CiMSC, 2023 [28]), stiefel manifold for multi-view clustering (SIREN, 2023 [10]).

**C. Parameters setting**: There are three parameters in the MVSC$^2$GF, namely, $\alpha, \beta$ and $\eta$. We search $\alpha, \beta$ in the interval $\{1e-5, 1e-4, 0.001, 0.01, 0.1, 0.2, 0.5, 0.8, 1, 2, 5, 8, 10\}$ and $\eta$ in $\{-5, -2, -1, 0.1, 0.5, 1.5, 2, 5\}$. The parameters in other evaluated algorithms are tuned based on the instructions in corresponding references.

### 4.2 Results and Analysis

The evaluated algorithms are performed on the benchmark multi-view data sets and the best results of the four metrics with parameters varying in corresponding intervals are recorded. These experiments are repeated 30 times. Then the means and standard deviations of the four metrics obtained by the evaluated algorithms are summarized in Table 1. The value of the parameters corresponding to the results of MVSC$^2$GF are listed in Table 2. From Table 1, we can get the following observations:

1. The proposed MVSC$^2$GF constantly outperforms other MVSC algorithms on all the data sets. Notably, on the 3sources, COIL-20, and Clatech101-7 data sets, MVSC$^2$GF achieves clustering accuracy that exceeds the second-best results by 3%. Specifically, on the BBC-sport database, all evaluation metrics of MVSC$^2$GF substantially outperform the second-best results by a considerable margin, approximately 10%. Moreover, the performance of MVSC$^2$GF is stable according to its standard deviations obtained on all data sets.

---
[1] The Matlab codes of MVSC$^2$GF will be released soon.
[2] http://mlg.ucd.ie/datasets/3sources.html
[3] https://cam-orl.co.uk/facedatabase.html
[4] http://mlg.ucd.ie/datasets/bbc.html
[5] https://www.cs.columbia.edu/CAVE/software/softlib/coil-20.php
[6] http://www.vision.caltech.edu/datasets/
[7] https://archive.ics.uci.edu/ml/datasets.php

**Table 1: Clustering results (in %) including mean and std (in the brackets) of evaluated methods on the used benchmark data sets. The best results are emphasized in red and bold, the second-best results are denoted in bold and italics.**

| Databases | Metrics | DiMSC | AGML | MLAN | mPAC | GMC | LMSC | LMVSC | LSRMVSC | CiMSC | SIREN | MVSC$^2$GF |
|---|---|---|---|---|---|---|---|---|---|---|---|---|
| 3sources | ACC | 79.53(0.3) | 76.43(0.4) | *81.70(1.2)* | 79.76(0.1) | 69.23(0.1) | 78.53(0.4) | 76.92(0.0) | 75.15(0.0) | 80.13(0.0) | 76.12(1.2) | **85.80(0.0)** |
|  | NMI | 71.91(0.9) | 69.32(0.0) | *75.43(0.0)* | 72.30(0.1) | 62.16(0.0) | 65.87(0.2) | 65.38(0.0) | 65.03(0.2) | 71.27(0.1) | 62.16(0.1) | **78.66(0.0)** |
|  | ARI | 63.71(0.2) | 60.43(0.2) | 66.64(0.0) | 64.52(0.0) | 65.56(0.2) | 50.56(0.5) | 60.57(0.3) | *69.64(0.2)* | 68.64(0.2) | 64.52(0.1) | **69.71(0.0)** |
|  | F-score | 67.43(0.2) | 72.43(0.1) | *82.46(0.4)* | 80.54(0.1) | 73.27(0.0) | 70.13(0.7) | 79.24(0.4) | 76.73(0.6) | 77.47(0.0) | 77.83(0.2) | **83.75(0.0)** |
| ORL | ACC | 79.70(2.6) | 72.93(2.0) | 68.50(0.0) | 71.31(1.4) | 63.25(0.0) | 82.75(1.2) | 71.33(2.3) | 79.00(0.0) | *89.56(2.8)* | 82.75(1.2) | **90.00(0.0)** |
|  | NMI | 90.45(1.3) | 89.14(1.0) | 83.12(0.0) | 86.79(0.0) | 85.71(0.0) | 92.81(0.5) | 84.19(2.5) | 92.39(0.0) | *94.22(0.2)* | 90.18(2.3) | **95.97(0.0)** |
|  | ARI | 73.76(3.3) | 54.79(5.0) | 33.16(0.0) | 51.23(0.0) | 33.67(0.0) | 75.90(0.0) | 69.77(0.1) | 77.43(2.5) | *83.56(2.3)* | 74.35(0.1) | **84.58(0.0)** |
|  | F-score | 74.39(3.25) | 70.34(3.0) | 33.47(1.1) | 68.74(3.2) | 35.90(0.0) | 77.53(0.1) | 73.12(0.0) | 76.52(0.0) | *84.30(0.4)* | 77.53(0.0) | **86.58(0.0)** |
| MSRC-v1 | ACC | 75.93(0.9) | 76.44(0.5) | 85.43(0.3) | 81.43(0.1) | *89.56(0.3)* | 77.02(2.2) | 88.10(0.2) | 68.67(1.3) | 78.30(3.0) | 88.18(0.8) | **91.90(0.0)** |
|  | NMI | 62.24(1.5) | 77.65(2.1) | 75.13(0.3) | 75.08(0.0) | *82.00(1.9)* | 67.95(2.5) | 81.11(2.6) | 55.63(0.0) | 76.09(2.0) | 78.49(0.9) | **85.37(0.0)** |
|  | ARI | 54.80(1.5) | 59.07(0.4) | 70.94(0.4) | 61.27(0.0) | *76.74(2.3)* | 59.88(2.8) | 72.79(0.3) | 46.42(0.1) | 70.51(5.0) | 74.73(0.8) | **82.57(0.0)** |
|  | F-score | 61.12(1.3) | 72.80(2.2) | 75.03(0.3) | 73.20(1.3) | 79.97(0.2) | 65.44(2.4) | 76.65(0.0) | 61.62(0.0) | 70.02(0.0) | *81.12(1.3)* | **85.02(0.0)** |
| BBCsport | ACC | *89.71(0.0)* | 69.34(0.4) | 79.62(0.0) | 77.96(0.0) | 80.63(0.0) | 85.12(12.3) | 48.78(3.4) | 80.98(2.5) | 72.05(0.1) | 79.74(3.5) | **98.16(0.0)** |
|  | NMI | *79.20(0.0)* | 56.39(0.1) | 77.91(0.0) | 73.30(0.1) | 75.68(0.0) | 74.48(13.6) | 27.21(1.6) | 84.93(0.0) | 68.39(8.0) | 76.43(2.5) | **93.27(0.0)** |
|  | ARI | *83.20(0.0)* | 70.48(0.5) | 65.54(0.0) | 68.48(2.0) | 72.20(0.0) | 77.04(16.5) | 31.45(1.4) | 81.34(0.2) | 60.04(1.0) | 71.30(1.1) | **95.03(0.0)** |
|  | F-score | *87.32(0.0)* | 73.54(0.8) | 72.44(0.0) | 82.04(0.0) | 79.41(0.0) | 82.88(11.4) | 39.34(5.6) | 86.03(0.0) | 68.13(0.0) | 80.13(0.0) | **96.22(0.0)** |
| COIL20 | ACC | 77.85(2.2) | 73.21(1.3) | *86.21(1.1)* | 72.43(0.0) | 79.11(0.1) | 77.97(0.1) | 70.83(0.3) | 70.56(1.2) | 85.98(0.0) | 76.54(1.2) | **90.00(0.0)** |
|  | NMI | 84.63(0.2) | 79.54(0.8) | *96.10(0.4)* | 84.46(0.0) | 94.10(0.2) | 85.92(0.1) | 81.49(0.2) | 79.97(0.0) | 94.56(3.2) | 81.34(0.1) | **97.10(0.0)** |
|  | ARI | 73.21(0.5) | 66.32(1.5) | *83.56(0.6)* | 71.44(0.0) | 78.20(0.0) | 67.52(0.2) | 59.41(0.6) | 62.89(0.3) | 82.32(0.2) | 74.56(0.5) | **84.34(0.0)** |
|  | F-score | 74.50(0.5) | 69.40(0.4) | *84.43(1.3)* | 73.02(0.0) | 80.21(0.0) | 71.61(0.2) | 63.95(0.5) | 64.98(0.0) | 81.65(0.0) | 77.38(0.0) | **86.80(0.0)** |
| Caltech101-7 | ACC | 51.74(0.1) | 68.73(0.3) | 62.70(0.0) | 59.65(0.0) | 67.52(2.2) | *86.33(0.1)* | 83.46(0.1) | 46.47(0.0) | 81.33(0.0) | 78.54(0.0) | **90.50(0.0)** |
|  | NMI | 31.73(0.7) | 49.43(11.5) | 54.67(0.0) | 46.24(0.0) | 61.99(1.9) | *72.97(0.4)* | 58.50(0.4) | 18.52(0.0) | 69.79(3.4) | 70.33(2.8) | **81.79(0.0)** |
|  | ARI | 38.63(0.6) | 53.21(0.0) | 41.47(0.0) | 51.23(2.0) | 41.17(0.4) | 41.17(0.4) | 38.04(0.5) | 13.00(0.0) | *54.33(7.3)* | 51.32(0.0) | **55.05(0.0)** |
|  | F-score | 42.11(0.8) | 61.53(6.9) | 61.88(0.0) | 57.51(0.0) | 68.29(0.0) | *91.28(0.2)* | 78.28(2.2) | 44.70(0.0) | 77.35(0.3) | 75.53(1.2) | **95.89(0.0)** |
| HW | ACC | 86.61(0.3) | 87.32(0.0) | 96.05(0.0) | 88.90(2.5) | 88.20(2.5) | 89.35(0.3) | 92.66(0.0) | 35.36(0.0) | 94.32(0.0) | *96.41(1.1)* | **96.59(0.0)** |
|  | NMI | 79.20(0.1) | 83.72(6.4) | 92.22(0.0) | 83.61(0.0) | 90.41(0.1) | 83.70(0.2) | 86.01(0.1) | 57.36(0.0) | 89.90(0.0) | *92.42(0.2)* | **92.59(0.0)** |
|  | ARI | 75.04(0.2) | 71.30(0.0) | 82.45(0.0) | 79.67(0.0) | 80.12(0.0) | 78.93(0.3) | 86.13(0.1) | 22.49(0.0) | 83.44(1.3) | *91.12(0.0)* | **91.65(0.0)** |
|  | F-score | 77.53(0.2) | 76.32(10.3) | 84.32(0.0) | 82.15(0.0) | 86.53(0.0) | 82.02(0.2) | 84.69(0.1) | 55.64(0.0) | 82.13(0.0) | *90.03(0.0)* | **92.48(0.0)** |

**Table 2: The value of the parameters corresponding to the results of MVSC$^2$GF in Table 1.**

| Databases | parameters | | |
|---|---|---|---|
|  | $\alpha$ | $\beta$ | $\eta$ |
| 3sources | 1 | 0.8 | 0.5 |
| ORL | 0.2 | 0.1 | 0.5 |
| MSRC-v1 | $1e-5$ | 0.5 | 0.5 |
| BBCsport | 0.2 | 2 | 0.5 |
| COIL20 | 0.5 | 0.1 | 0.5 |
| Caltech101-7 | 5 | 10 | 0.5 |
| HW | 0.8 | 0.5 | 0.5 |

2. MVSC$^2$GF demonstrates substantial superiority over LSR-MVSC which is also a graph filtering-based MVSC algorithm. Across all datasets, all the metrics obtained by MVSC$^2$GF surpass those of LSRMVSC by more than 10%. Especially, on Caltech101-7 and HW databases, LSRMVSC can not achieve satisfactory results while MVSC$^2$GF still achieves the best results. The reasons can be explained from two facts, namely the filtered features and the acquired reconstruction coefficient matrices in different views. Taking the BBCsport data set as an example, Fig. 2 and 3 display the filtered features obtained by the two algorithms in different views, along with the respective reconstruction coefficient matrices. These coefficient matrices correspond to the best experimental results achieved by the two algorithms. And the consensus matrix obtained by MVSC$^2$GF is also presented. Here, by following the Matlab codes of LSRMVSC [8], the two parameters in LSRMVSC are set to $\eta = 1000, \gamma = 80$, and the filter order is designated as 3. The three parameters in MVSC$^2$GF are set as $\alpha = 0.2, \beta = 2, \eta = 0.5$. From Fig. 2, we observe that though the features acquired through LSRMVSC generally display distinction among different classes, there exists an overlap among feature vectors belonging to different classes. On the other hand, the features obtained by MVSC$^2$GF exhibit minimal overlap across different classes. It should be pointed out that LSRMVSC gets a distinct graph filter in each view, so the filtered features are obtained by using the data and corresponding graph filter in each view. MVSC$^2$GF constructs a consensus graph filter, so the filtered features are computed by using the consensus graph filter and the data in each view. Additionally, as depicted in Fig. 3, the reconstruction coefficient matrices obtained by LSRMVSC and MVSC2GF all illustrate block-diagonal structures. But the coefficient matrices obtained by MVSC$^2$GF are more sparse than those of LSRMVSC, namely in the coefficient matrices obtained by MVSC$^2$GF, the coefficients corresponding to the samples from different classes are close to zero (Zoom in for a better view). Moreover, we can see the consensus coefficient matrix of MVSC$^2$GF shows a clearer block-diagonal structure than that of the coefficient matrix obtained in individual view. Hence, it is reasonable to use the consensus graph filter constructed by the consensus reconstruction coefficient matrix for feature smoothing.

[8]https://github.com/huangsd/LSR_MSC

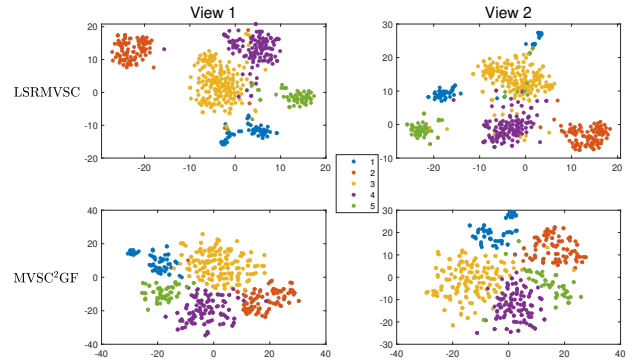

**Figure 2: The filtered features in each view obtained by LSR-MVSC (the first row) and MVSC$^2$GF (the second row). The filtered features from the first and second views are listed in the first and second columns respectively.**

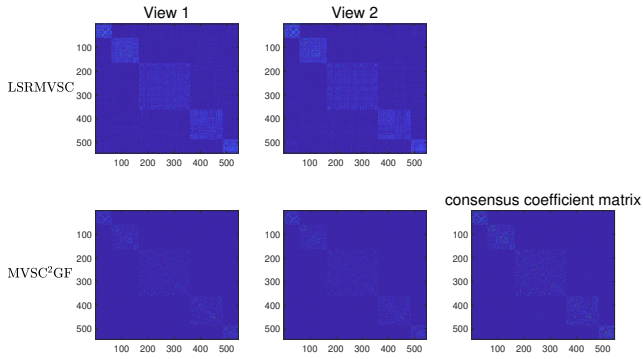

Figure 3: The reconstruction coefficient matrix in each view obtained by LSRMVSC (the first row) and MVSC$^2$GF (the second row). The coefficient matrix obtained in the first and second views are listed in the first and second columns respectively. The consensus coefficient matrix obtained by MVSC$^2$GF is visualized in the last figure of the second row.

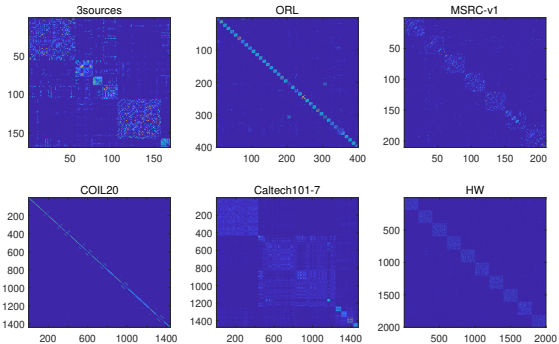

Figure 4: The consensus reconstruction coefficient matrices of MVSC$^2$GF obtained on the rest six data sets.

Additionally, we present the consensus coefficient matrices calculated by MVSC$^2$GF on the remaining six data sets in Fig. 4. These consensus coefficient matrices correspond to the optimal experimental results listed in Table 1. We can see that the obtained consensus coefficient matrix is capable of capturing the underlying subspace structure within the associated data set.

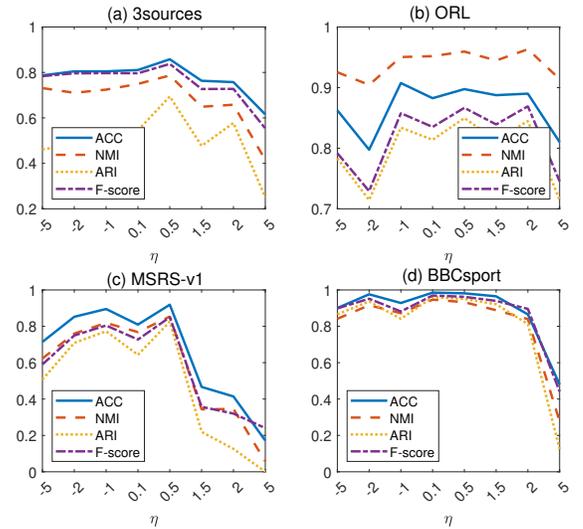

Figure 5: The performance of MVSC$^2$GF changes with $\eta$.

### 4.3 Parameter analysis

There are three parameters in MVSC$^2$GF. $\alpha, \beta$ are used to balance the effects of two terms in the objection of MVSC$^2$GF, and $\eta$ is used to adjust the importance of each view. We first observe the impact of $\eta$ on the MVSC$^2$GF. Here, the four data sets including 3sources, ORL, MSRC-v1, and BBCsport are used. We vary the parameter $\eta$ across the range $\{-5, -2, -1, 0.1, 0.5, 1.5, 2, 5\}$. Maintaining a fixed $\eta$, we record the best results achieved by MVSC$^2$GF while altering $\alpha$ and $\beta$ vary in $\{1e-5, 1e-4, 0.001, 0.01, 0.1, 0.2, 0.5, 0.8, 1, 2, 5, 8, 10\}$. The resulting curves, which present the four evaluation metrics against $\eta$ on the four data sets, are represented in Fig. 5. We can see when $\eta \leq 0.5$, MVSC$^2$GF can achieve good results on all the data sets.

Then we demonstrate how the performance of MVSC$^2$GF is affected by $\alpha$ and $\beta$. Based on the previous experiments, we fix $\eta = 0.5$, then the NMI variations of MVSC$^2$GF with respect to $\alpha, \beta$ across all data sets are depicted in Fig. 6. The figure indicates that the performance of MVSC$^2$GF remains consistent within a moderately large interval as $\alpha$ and $\beta$ change. The other three metrics show similar characteristics.

### 4.4 Convergence analysis

We now show the convergence of Algorithm 1. On each data set, we record the computed consensus coefficient matrix $\mathbf{C}$ and its corresponding auxiliary variable $\mathbf{Z}$ in each iteration of Algorithm 1. Then the residuals of the two variables (i.e. $\|\mathbf{C}_{k+1} - \mathbf{C}_k\|_F^2$ and $\|\mathbf{Z}_{k+1} - \mathbf{Z}_k\|_F^2$) can be computed, where $k$ denotes the number of iterations. The residuals versus the number of iterations are illustrated in Fig. 7. We can see Algorithm 1 can help MVSC$^2$GF problem find a stable set of solutions on all the data sets after fewer than 500 iterations.

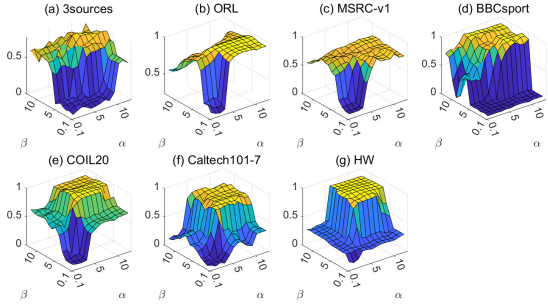

Figure 6: The NMI varies with $\alpha, \beta$ on the evaluated data sets.

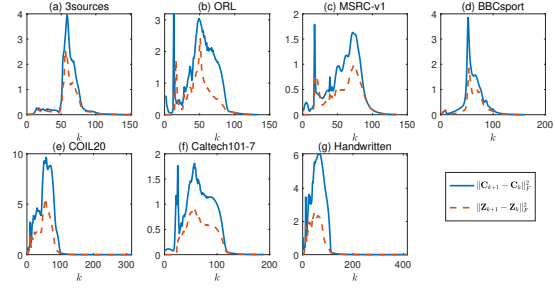

Figure 7: The residuals of variables C, Z versus the iterations on the evaluated databases. The horizontal coordinate represents the number of iterations.

### 4.5 Ablation study

The main innovations in MVSC$^2$GF involve two key aspects: smoothing features using a consensus graph filter in each view, and designing a regularizer for the reconstruction coefficient matrix in that view using the consensus filter. We discuss the importance of the two operations. Firstly, if we do not perform data smoothing, the objective function Problem (5) can be transformed as follows:

$$\min_{\mathbf{C}^i,\mathbf{Z}^i,\mathbf{C},\mathbf{Z},\gamma^i} \quad \sum_{i=1}^{v} \|\mathbf{X}^i - \mathbf{C}^i\mathbf{X}^i\|_F^2 + \alpha\|\mathbf{C}^i - \mathbf{C}\mathbf{C}^i\|_F^2 \\ + \beta(\gamma^i)^\eta\|\mathbf{C} - \mathbf{C}^i\|_F^2, \quad (22)$$

On the other hand, we replace the new regularizer with a Frobenius norm regularizer of $\mathbf{C}^i$, the following objective function can be gotten:

$$\min_{\mathbf{Y}^i,\mathbf{C}^i,\mathbf{Z}^i,\mathbf{C},\mathbf{Z},\gamma^i} \quad \sum_{i=1}^{v} \|\mathbf{Y}^i - \mathbf{C}^i\mathbf{Y}^i\|_F^2 + \alpha\|\mathbf{C}^i\|_F^2 \\ + \beta(\gamma^i)^\eta\|\mathbf{C} - \mathbf{C}^i\|_F^2, \quad (23)$$

By integrating the corresponding constraints in Problem (5), the above two problems can be solved. We record the results obtained by the two problems in Table 3. The results of MSVC$^2$GF are repeated in this table for clear comparison. We can see the performance of Problem (22) and (23) are both inferior to those of MSVC$^2$GF which proves the necessity of the proposed techniques in MSVC$^2$GF. Moreover, the experimental results of Problem (23) are better than those of Problem (22), which suggests that feature smoothing plays a more important role.

Table 3: Clustering results (mean in %) of Problem (22), (23) and MSVC$^2$GF on the used benchmark data sets. The best results are emphasized in red and bold.

| Databases | Metrics | Problem (22) | Problem (23) | MSVC$^2$GF |
|---|---|---|---|---|
| 3sources | ACC | 66.86 | 79.29 | **85.80** |
| | NMI | 57.37 | 70.58 | **78.66** |
| | ARI | 25.96 | 53.51 | **69.41** |
| | F-score | 64.05 | 78.71 | **83.75** |
| ORL | ACC | 83.75 | 65.75 | **90.00** |
| | NMI | 92.02 | 76.86 | **95.97** |
| | ARI | 75.79 | 49.04 | **84.58** |
| | F-score | 75.56 | 49.64 | **86.58** |
| MSRc-v1 | ACC | 59.05 | 90.03 | **91.90** |
| | NMI | 50.00 | 81.07 | **85.37** |
| | ARI | 34.04 | 79.65 | **82.57** |
| | F-score | 46.73 | 80.02 | **85.02** |
| BBCsport | ACC | 40.62 | 95.22 | **98.16** |
| | NMI | 9.89 | 86.31 | **93.27** |
| | ARI | 2.59 | 87.68 | **95.03** |
| | F-score | 39.23 | 90.70 | **96.22** |
| COIL20 | ACC | 73.82 | 51.32 | **90.00** |
| | NMI | 82.52 | 59.79 | **97.17** |
| | ARI | 65.31 | 34.43 | **84.34** |
| | F-score | 68.09 | 41.14 | **86.80** |
| Caltech101-7 | ACC | 56.95 | 83.31 | **90.50** |
| | NMI | 4.43 | 62.59 | **81.79** |
| | ARI | 4.17 | 30.23 | **55.05** |
| | F-score | 52.15 | 86.54 | **95.89** |
| HW | ACC | 14.50 | 36.65 | **96.59** |
| | NMI | 1.36 | 26.51 | **92.59** |
| | ARI | 0.26 | 15.37 | **91.65** |
| | F-score | 15.06 | 24.32 | **92.48** |

## 5 CONCLUSIONS

In this paper, we introduce a new multi-view subspace clustering method, termed multi-view subspace clustering by an adaptive consensus graph filter (MVSC$^2$GF). MVSC$^2$GF assumes the existence of a consensus reconstruction coefficient matrix, leveraging it to construct a consensus graph filter. This filter is employed to smooth data features across different views while designing a regularizer for the reconstruction coefficient matrix in each view. Then the obtained reconstruction coefficient matrices from different views can form constraints for the consensus reconstruction coefficient matrix. Due to the interdependency of the consensus reconstruction coefficient matrix and reconstruction coefficient matrices across different views, we present an optimization algorithm to iteratively optimize these variables. We also show the convergence of the optimized algorithm. Through extensive subspace clustering experiments, we demonstrate that our proposed method achieves state-of-the-art (SOTA) performance.